\documentclass[sigconf]{acmart}

\AtBeginDocument{%
  \providecommand\BibTeX{{%
    \normalfont B\kern-0.5em{\scshape i\kern-0.25em b}\kern-0.8em\TeX}}}

\setcopyright{rightsretained}

\usepackage{algpseudocode}
\usepackage{enumitem}
\usepackage{algorithm}
\usepackage{graphicx}
\usepackage{subcaption}

\newcommand{\x}{{\mathbf{x}}}

\newcommand{\train}{{tr}}
\newcommand{\valid}{{va}}
\newcommand{\test}{{te}}

\newcommand{\beq}{\begin{equation}}
\newcommand{\eeq}{\end{equation}}
\newcommand{\bea}{\begin{eqnarray}}
\newcommand{\eea}{\end{eqnarray}}
\newcommand{\beas}{\begin{eqnarray*}}
\newcommand{\eeas}{\end{eqnarray*}}
\newcommand{\ea}{\end{array}}
\newcommand{\bit}{\begin{itemize}}
\newcommand{\eit}{\end{itemize}}
\newcommand{\ben}{\begin{enumerate}}
\newcommand{\een}{\end{enumerate}}
\newcommand{\bde}{\begin{description}}
\newcommand{\ede}{\end{description}}
\newcommand{\bsp}{\begin{split}}
\newcommand{\esp}{\end{split}}

\newcommand{\monthyear}{%
  \ifcase\month\or January\or February\or March\or April\or May\or June\or
  July\or August\or September\or October\or November\or
  December\fi\space\number\year
}

\usepackage{cleveref}

\hypersetup{draft}

\begin{document}

\title{Meta-Learning for Few-Shot Time Series Classification}

\author{ Jyoti Narwariya, Pankaj Malhotra, Lovekesh Vig, Gautam Shroff, Vishnu TV}
\email{ jyoti.narwariya@tcs.com, malhotra.pankaj@tcs.com, lovekesh.vig@tcs.com, gautam.shroff@tcs.com, vishnu.tv@tcs.com}
\affiliation{
	\institution{TCS Research}
	\city{New Delhi, India}
}

\begin{abstract}
Deep neural networks (DNNs) have achieved state-of-the-art results on time series classification (TSC) tasks. 
In this work, we focus on leveraging DNNs in the often-encountered practical scenario where access to labeled training data is difficult, and where DNNs would be prone to overfitting.
We leverage recent advancements in gradient-based meta-learning, and propose an approach to train a residual neural network with convolutional layers as a meta-learning agent for few-shot TSC.
The network is trained on a diverse set of few-shot tasks sampled from various domains (e.g. healthcare, activity recognition, etc.) such that it can solve a target task from another domain using only a small number of training samples from the target task.
Most existing meta-learning approaches are limited in practice as they assume a fixed number of target classes across tasks.
We overcome this limitation in order to train a common agent across domains with each domain having different number of target classes, we utilize a triplet-loss based learning procedure that does not require any constraints to be enforced on the number of classes for the few-shot TSC tasks.
To the best of our knowledge, we are the first to use meta-learning based pre-training for TSC. 
Our approach sets a new benchmark for few-shot TSC, outperforming
several strong baselines on few-shot tasks sampled from 41 datasets in UCR TSC Archive.
We observe that pre-training under the meta-learning paradigm allows the network to quickly adapt to new unseen tasks with small number of labeled instances.
\end{abstract}


\maketitle

\section{Introduction}
Time series data is ubiquitous in the current digital era with several applications across domains such as forecasting, healthcare, equipment health monitoring, and meteorology among others.
Time series classification (TSC) has several practical applications such as disease diagnosis from time series of physiological parameters \cite{che2018recurrent}, classifying heart arrhythmias from ECG signals\cite{rajpurkar2017cardiologist}, and human activity recognition \cite{yang2015deep}.
Recently, deep neural networks (DNNs) such as those based on long short term memory networks (LSTMs) \cite{karim2018lstm} and 1-dimensional convolution neural networks (CNNs) \cite{wang2017time,fawaz2018deep,kashi2019convtimenet} have achieved state-of-the-art results on TSC tasks.
However, it is well-known that DNNs are prone to overfitting, especially when access to a large labeled training dataset is not available. \cite{fawaz2018transfer,kashi2019convtimenet}.

Few recent attempts aim to address the issue of scarce labeled data for univariate TSC (UTSC) by leveraging transfer learning \cite{yosinski2014transferable} via DNNs, e.g. \cite{malhotra2017timenet,serra2018towards,fawaz2018transfer,kashi2019convtimenet}.
These approaches consider pre-training a deep network in an unsupervised \cite{malhotra2017timenet} or supervised \cite{serra2018towards,fawaz2018transfer,kashi2019convtimenet} manner using a large number of time series from diverse domains, and then fine-tune the pre-trained model for the target task using labeled data from target domain.
However, these transfer learning approaches for TSC based on pre-training a network on large number of diverse time series tasks do not necessarily guarantee a pre-trained model (or network initialization) that can be quickly fine-tuned with a very small number of labeled training instances, and rather rely on ad-hoc fine-tuning procedures.

Rather than learning a new task from scratch, humans leverage their pre-existing skills by fine-tuning and recombining them, and hence are highly data-efficient, i.e. can learn from as little as one example per category \cite{perfors2009learning}. 
Meta-learning \cite{schmidhuber1987evolutionary} approaches intend to take a similar approach for \textit{few-shot learning}, i.e. learning a task from few examples. 
More recently, several approaches for few-shot learning for regression, image classification, and reinforcement learning domains have been proposed under the gradient-based meta-learning or the ``learning to learn" framework, e.g. in \cite{finn2017model,nichol2018first,rusu2018meta}.
A neural network-based meta-learning model is explicitly trained to quickly learn a new task from a small amount of data. 
The model learns to solve several tasks sampled from a given distribution where each task is, for example, an image classification problem with few labeled examples.
Since each task corresponds to a learning problem, performing well on a task corresponds to learning quickly.

Despite the advent of aforementioned pre-trained models for time series, \textit{few-shot learning} (i.e. learning from few, say five, examples per class) for TSC remains an important and unaddressed research problem.
The goal of few-shot TSC is to train a model on large number of diverse few-shot TSC tasks such that it can leverage this experience through the learned parameters, and quickly generalize to new tasks with small number of labeled instances.
More specifically, we train a residual network (ResNet) \cite{wang2017time,fawaz2018deep} on several few-shot TSC tasks such that the ResNet thus obtained generalizes to solve new few-shot learning tasks.
In contrast to existing methods for data-efficient transfer learning, our method provides a way to directly optimize the embedding itself for classification, rather than an intermediate bottleneck layer such as the ones proposed in \cite{malhotra2017timenet,serra2018towards}.

Key contributions of this work are: 
\begin{itemize}
	\item We define the problem of few-shot learning for univariate TSC (UTSC), and propose a training and evaluation protocol for the same. 
	\item We propose a few-shot UTSC approach by training a ResNet to solve diverse few-shot UTSC tasks using a meta-learning procedure \cite{nichol2018first}. The ResNet thus obtained can be quickly adjusted (fine-tuned) on a new, previously unseen, UTSC task with few labeled examples per class.
	\item As opposed to fixed $N$-way classification setting in most existing few-shot methods, our approach can handle multi-way classification problems with varying number of classes without introducing any additional task-specific parameters to be trained from scratch such as those in the final classification layer \cite{serra2018towards,kashi2019convtimenet}: In order to generalize across few-shot tasks with varying number of classes, we leverage triplet loss \cite{weinberger2006distance,schroff2015facenet} for training the ResNet. This allows our approach to leverage the same neural network architecture across diverse applications without introducing any additional task-specific parameters to be trained from scratch.
	\item Since the proposed approach uses triplet loss to learn a Euclidean embedding for time series, it can also be seen as a data-efficient metric learning procedure for time series that can learn from very small number of labeled instances.
\end{itemize}

In few-shot setting, we demonstrate that a vanilla nearest-neighbor classifier over the embeddings obtained using our approach outperforms existing nearest-neighbor classifiers based on the highly effective dynamic time warping (DTW) classifier \cite{bagnall2017great} and even state-of-the-art time series classifier BOSS \cite{schafer2015boss}. 
The rest of the paper is organized as follows: we contrast our work to existing literature in Section \ref{sec:rw}. We define the problem of few-shot learning for UTSC in Section \ref{sec:pd}. We then provide details of the neural network architecture used for training the few-shot learner in Section \ref{sec:fcn} followed by the details of meta-learning based training algorithm for few-shot UTSC in Section \ref{sec:fsl}. We provide details of empirical evaluation of proposed approach in Section \ref{sec:exp} and conclude in Section \ref{sec:conc}.

\section{Related Work\label{sec:rw}}
Several approaches have been proposed to deal with scarce labeled data for TSC, via data augmentation, warping, simulation, transfer learning, etc. in e.g. \cite{le2016data,cui2016multi,fawaz2018data,kashi2019convtimenet}. 
Regularization in DNNs, e.g. decorrelating convolutional filter weights \cite{kaushal2019regularizing} has been found to be effective for TSC and avoid overfitting in scarce data scenarios.
Iterative semi-supervised learning \cite{wei2006semi} also addresses scarce labeled data scenario by iteratively increasing the labeled set but assumes availability of a relatively large amount of data albeit initially unlabeled.
In this work, we take a different route to deal with scarce labeled data scenarios and leverage gradient-based meta-learning to explicitly train a network to quickly adapt and solve new few-shot TSC tasks. 

Transfer learning using pre-trained DNNs has been shown to achieve better classification performance than training DNNs from scratch for TSC: a few instances of pre-trained DNNs for TSC have been recently proposed in e.g. \cite{malhotra2017timenet,serra2018towards,kashi2019convtimenet}. 
However, none of these methods are explicitly trained to quickly adapt to a target task and tend to rely on ad-hoc fine-tuning procedures. 
Furthermore, they do not study the extreme case of few-shot TSC: while \cite{malhotra2017timenet} relies on training an SVM classifier on top of unsupervised embeddings obtained via a deep LSTM, \cite{serra2018towards,kashi2019convtimenet} rely on introducing and training a new final softmax layer from scratch for each new task. 
Our approach explicitly pre-trains a DNN using triplet loss to optimize for quick adaptation to a few-shot task. 
Moreover, unlike existing methods, our approach directly optimizes for time series embeddings over which the similarity of time series can be defined, and hence can work in a kNN setting without requiring the training of additional parameters like those of an SVM in \cite{malhotra2017timenet}, or those of a feedforward final layer in \cite{serra2018towards,kashi2019convtimenet}. 
 
Several approaches for few-shot learning have been recently introduced for image classification, regression, and reinforcement learning, e.g. \cite{vinyals2016matching,finn2017model,nichol2018first,rusu2018meta}.
To the best of our knowledge, our work is the first attempt to study few-shot learning for TSC.
We formulate the few-shot learning problem for UTSC, and build on top of the following recent advances in deep learning research to develop an effective few-shot approach for TSC: 
i) gradient-based meta-learning \cite{finn2017model,nichol2018first}, 
ii) residual network with convolutional layers for TSC \cite{wang2017time}, 
iii) leveraging multi-length filters to ensure generalizability of filters to tasks with varying time series length and temporal properties \cite{roy2018chrononet,kashi2019convtimenet}, and 
iv) triplet loss \cite{schroff2015facenet} to ensure generalizability to tasks with varying number of classes without introducing any additional parameters.

Dynamic time warping (DTW) and its variants \cite{vintsyuk1968speech,jeong2011weighted} are known to be very robust and strong distance metric baselines for TSC over a diverse set of applications \cite{bagnall2017great}. 
However, it is also well-known that no single distance metric works well across scenarios as they lack the ability to leverage the data-distribution and properties of the task at hand \cite{wang2013effectiveness,bagnall2017great}. 
It has been shown that k-nearest-neighbor (kNN) TSC can be significantly improved by learning a distance metric from labeled examples \cite{mei2015learning,abid2018learning}.
Similarly, modeling time series similarity using Siamese recurrent networks based supervised learning has been proposed in \cite{pei2016modeling}.
CNNs trained using triplet loss for TSC have been very recently proposed for unsupervised learning in \cite{franceschi2019unsupervised} and for supervised learning in \cite{brunel2019cnn}.
However, to the best of our knowledge, none of the metric learning approaches consider pre-training a neural network that can be quickly fine-tuned for new TSC few-shot tasks. 

\section{Problem Definition\label{sec:pd}}
Consider a $K$-shot learning problem for UTSC sampled from a distribution $p(\mathcal{T})$ that requires learning a multi-way classifier for a test task given only $K$ labeled time series instances per class.
Rather than training a classifier from scratch for the test task, the goal is to obtain a neural network with parameters $\phi$ that is trained to efficiently (e.g. in a few iterations of updates of $\phi$ via gradient descent) solve several $K$-shot learning tasks sampled from $p(\mathcal{T})$.
These $K$-shot tasks are divided into three sets: a \textit{training meta-set} $\mathcal{S}^{\train}$, a \textit{validation meta-set} $\mathcal{S}^{\valid}$, and a \textit{testing meta-set} $\mathcal{S}^{\test}$.
The training meta-set is used to obtain the parameters $\phi$, the validation meta-set is used for model selection (hyperparameters for neural network training), and the testing meta-set is used only for final evaluation.

Each task instance $\mathcal{T}_{j} \sim p\left(\mathcal{T}\right)$ in $\mathcal{S}^{\train}$ and $\mathcal{S}^{\valid}$ consists of a labeled training set of univariate time series $\mathcal{D}^{\train}_j = \big\{(\x_j^{n,k}, y_j^{n,k}) \mid k = 1 \dots K ; n = 1 \dots N_j \big\}$, where $K$ is the number of univariate time series instances for each of the $N_j$ classes. 
Ignoring the sub- and super-scripts, each univariate time series $\x = x_1,x_2,\dots,x_{T}$ with $x_t \in \mathbb{R}$ for $t=1,\dots,T$, where $T$ is the length of time series, and $y$ is the class label. 
Unlike the tasks in $\mathcal{S}^{\train}$ and $\mathcal{S}^{\valid}$, which only contain a training set, each task in $\mathcal{S}^{te}$ also contains a testing set  $\mathcal{D}^{\test}_j = \big\{(\x_j^{n,k}, y_j^{n,k}) \mid k = 1 \dots K' ; n = 1 \dots N_j \big\}$ apart from a training set $\mathcal{D}^{\train}_j$. 
The classes in $\mathcal{D}^{\train}_j$ and $\mathcal{D}^{\test}_j$ are the same while classes across tasks are, in general, different.
For any $\x_j^{n,k}$ from $\mathcal{D}^{\test}_j$, the goal is to estimate the corresponding label $y_j^{n,k}$ by using an updated version $\tilde\phi$ of $\phi$ obtained by fine-tuning the neural network using the $K\times N_j$ labeled samples from $\mathcal{D}^{\train}_j$.
In other words, the training set $\mathcal{D}^{\train}_j$ of a task $\mathcal{T}_j \in \mathcal{S}^{\test}$ is used for fine-tuning the neural network parameters $\phi$, while the corresponding testing set $\mathcal{D}^{\test}_j$ of the task $\mathcal{T}_j$ is used for evaluation.

It is to be noted that the tasks in the three meta-sets correspond to time series from disjoint sets of classes, i.e. the classes in any task in training meta-set are different from those of any task in validation meta-set, and so on.
In practice, we sample the tasks from diverse domains such as electric devices,
motion capture, spectrographs, sensor readings, ECGs, simulated time series, etc. taken from the UCR TSC Archive \cite{UCRArchive}. Each dataset, and in turn tasks sampled from it, have a different notion of classes depending upon the domain, a different number of classes $N$, and a different  $T$. 

\section{Neural Network \label{sec:fcn}}

\begin{figure}[h]
	\centering
	\includegraphics[scale=0.35, angle=-180,trim={1cm, 1.5cm, 1cm, 1.25cm},clip]{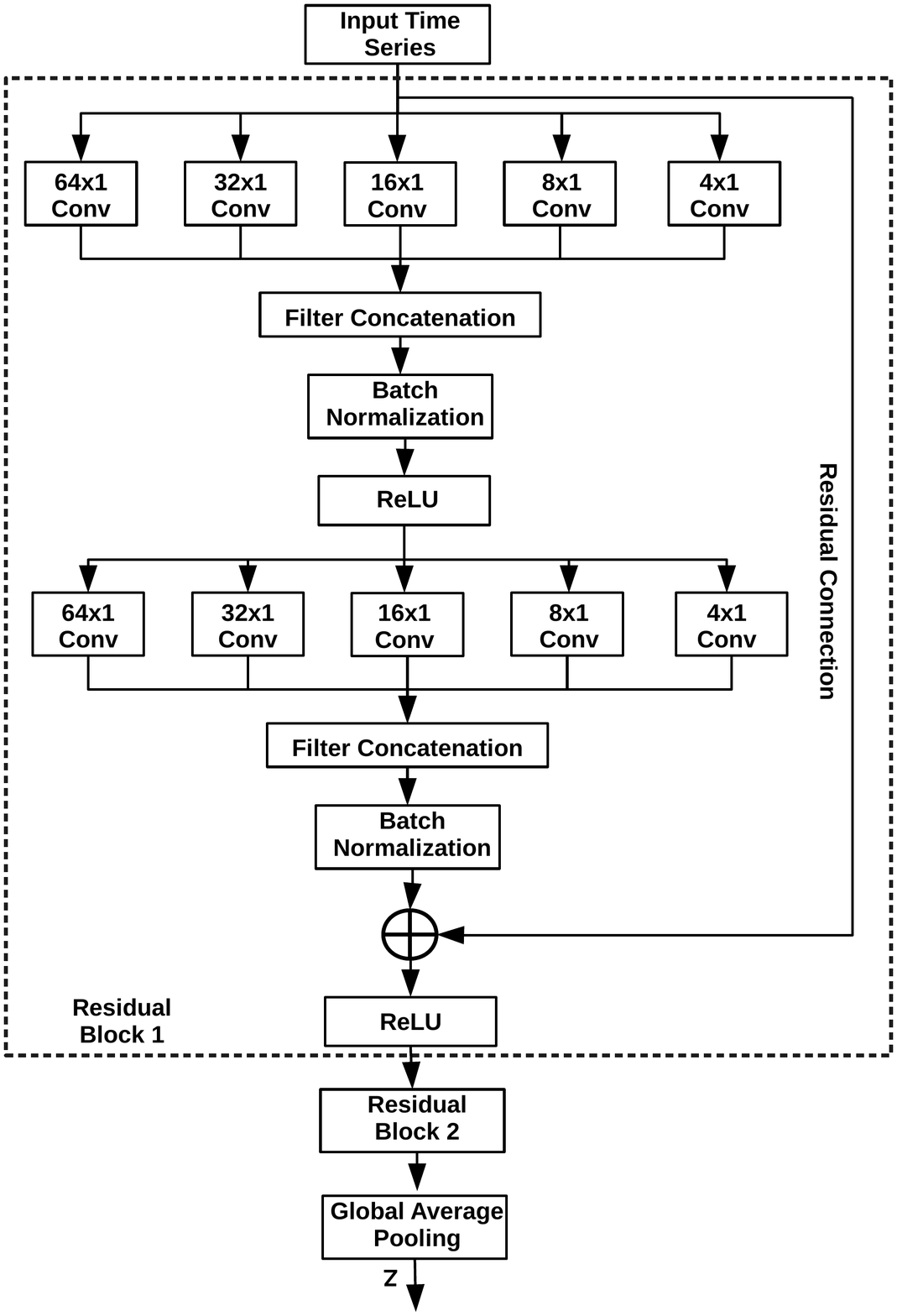}
	\caption{ResNet Architecture depicting two residual blocks each with two convolutional layers, and variable-length filters in each convolutional layer.}
	\label{fig:fcn}
\end{figure}

As shown in Figure \ref{fig:fcn}, we consider a ResNet consisting of multiple convolutional blocks with shortcut residual connections \cite{he2016deep} between them, eventually followed by a global average pooling (GAP) layer such that the network does not have any feedforward layers at the end.
Each convolutional block consists of a convolutional layer followed by a batch normalization (BN) layer \cite{ioffe2015batch} which acts as a regularizer. Each BN layer is in turn followed by a ReLU layer. 
We omit further architecture details and refer the reader to \cite{kashi2019convtimenet}.
In order to quickly adapt to any unseen task, the neural network should be able to extract temporal features at multiple time scales and should ensure that the fine-tuned network can generalize to time series of varying lengths across tasks. 
We, therefore, use filters of multiple lengths in each convolutional block to capture temporal features at various time scales, as found to be useful in \cite{roy2018chrononet,brunel2019cnn,kashi2019convtimenet}. 

In a nutshell, ResNet takes a univariate time series $\mathbf{x}$ of any length $T$ as input and converts it to a fixed-dimensional feature vector $\mathbf{z} \in \mathbb{R}^m$, where $m$ is the number of filters in the final convolutional layer.
We denote the set of all the trainable parameters of the ResNet consisting of filter weights and biases across convolutional layers, and BN layer parameters by $\phi$.

Most ResNet implementations for TSC \cite{wang2017time,fawaz2018deep,serra2018towards,kashi2019convtimenet} use a feedforward layer followed by a softmax layer to eventually map $\mathbf{z}$ to class probabilities, and use cross-entropy loss for training. 
Further, when training the ResNet for multiple tasks with varying number of classes across tasks, a multi-head output with different final feedforward layer for each task is typically used, e.g. as in \cite{serra2018towards,kashi2019convtimenet}.
However, in our setting, this implies a different feedforward layer for each new few-shot task, introducing at least $m\times N_j$ additional task-specific parameters\footnote{when the GAP layer is followed by a single feedforward layer and a softmax layer} that need to be trained from scratch for each new few-shot task.
This is not desirable in a few-shot learning setting given only a small number of $K$ samples per class, as this can lead to overfitting: this is one reason due to which most few-shot learning formulations, e.g. \cite{vinyals2016matching,finn2017model}, consider a fixed number of target classes across tasks.
However, we intend to learn a few-shot learning algorithm that overcomes this limitation.
We propose using triplet loss \cite{weinberger2006distance,schroff2015facenet,brunel2019cnn} as the training objective which allows for generalization to varying number of classes without introducing any additional task-specific parameters, as detailed next.

\subsection{Loss Function\label{ssec:loss}}
Triplet loss relies on pairwise distance between representations of time series samples from within and across classes, irrespective of the number of classes.
Using triplet loss at time of fine-tuning for the test task, therefore, allows the network to adapt to a given few-shot classification task without introducing any additional task-specific parameters. 
Triplets consist of two matching time series and a non-matching time series such that the loss aims to separate the positive pair from the negative by a distance margin.
Given the set $\mathcal{S}_j$ of all valid triplets of time series for a training task $\mathcal{T}_j$ of the form $\left(\mathbf{x}_l^a, \mathbf{x}_l^p, \mathbf{x}_l^n\right)\in\mathcal{S}_j$ consisting of an anchor time series $\mathbf{x}_l^a$, a positive time series $\mathbf{x}_l^p$, and a negative time series $\mathbf{x}_l^n$; where the positive time series is another instance from same class as the anchor, while the negative is from a different class than the anchor.
We aim to obtain corresponding representations $\left(\mathbf{z}_l^a, \mathbf{z}_l^p, \mathbf{z}_l^n\right)$ such that the distance between the representations of an anchor and any positive time series is lower than the distance between the representations of the anchor and any negative time series. 

More specifically, we consider triplet loss based on Euclidean norm given by: 
\begin{equation}\label{eq:triplet_constraint}
\|\mathbf{z}_l^a - \mathbf{z}_l^n\|_2^2 - \|\mathbf{z}_l^a - \mathbf{z}_l^p\|_2^2 > \alpha \;,\\
\end{equation}
where $\alpha>0$ is the distance-margin between the positive and negative pairs.
The loss to be minimized is then given by:
\begin{equation}\label{eq:triplet_loss}
\mathcal{L}_{\mathcal{T}_j} = \sum_{l=1}^{|\mathcal{S}_j|}\left[\|\mathbf{z}_l^a-\mathbf{z}_l^p\|_2^2 -
\|\mathbf{z}_l^a-\mathbf{z}_l^n\|_2^2+\alpha\right]_+,
\end{equation}
where $[z]_+=max(z,0)$, such that only those triplets violating the constraint in Eq. \ref{eq:triplet_constraint} contribute to the loss. Note that since we use triplet loss for training, the number of instances per class $K>1$.

\section{Few-Shot Learning for UTSC \label{sec:fsl}}
\begin{figure}[h]
	\centering
	\includegraphics[width=\columnwidth,trim={1cm 3.5cm 1cm 1.5cm},clip]{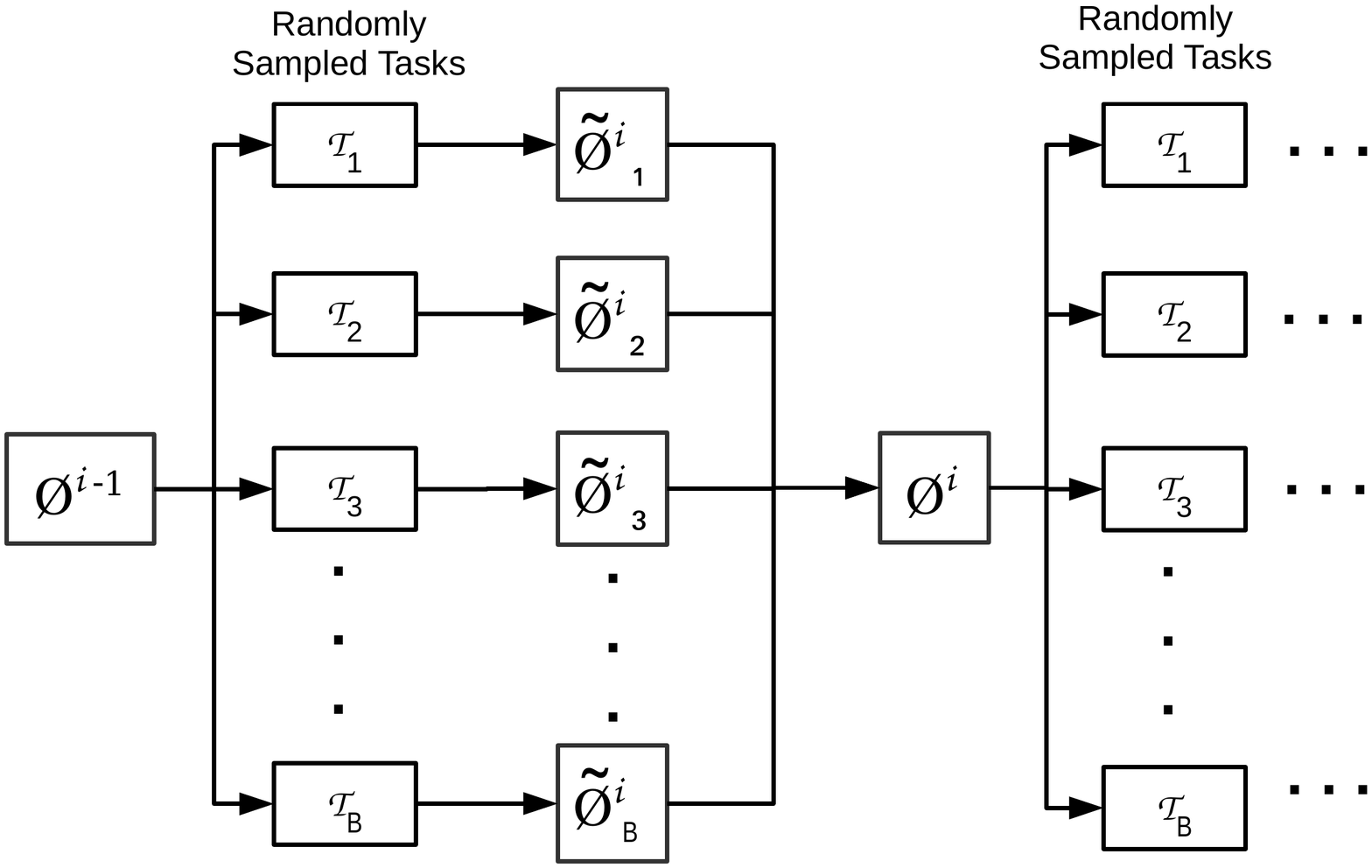}
	\caption{Few-Shot Training Approach.}
	\label{fig:training}
\end{figure}
We consider a meta-learning approach for few-shot UTSC based on Reptile \cite{nichol2018first}, a first-order gradient descent based meta-learning algorithm, and refer to that as \textit{FS-1}. 
We also consider a simpler variant of this approach and refer to that as \textit{FS-2}: similar to the training procedure of FS-1, FS-2 is also trained to solve multiple UTSC tasks but not explicitly trained in a manner that ensures quick adaptation to any new UTSC task.
Except for the triplet loss, FS-2 is similar to \cite{serra2018towards,kashi2019convtimenet} in the way data is sampled and used for training.

\subsection{FS-1\label{ssec:ml}}

\subsubsection{\textbf{Objective}}
FS-1 learns an initialization for the parameters $\phi$ of the ResNet such that these parameters can be quickly optimized using gradient-based learning at test time to solve a new few-shot UTSC task---i.e., the model generalizes from a small number of examples from the test task.
In order to learn the parameters $\phi$, we train the ResNet on a diverse set of UTSC tasks in $\mathcal{S}^\train$ with varying number of classes and time series lengths.
As explained in Section \ref{sec:fcn}, the same neural network parameters $\phi$ are shared across all tasks owing to the fact that: i. ResNet yields a fixed-dimensional representation for varying length time series, and ii. the nature of the loss function that does not require any changes due to the varying number of classes across tasks.

Similar to \cite{finn2017model,nichol2018first}, we consider the following optimization problem: find an initial set of parameters $\phi$ for the ResNet, such that for a randomly sampled task $\mathcal{T}_j$ with corresponding loss $L_{\mathcal{T}_j}$ as given in Eq. \ref{eq:triplet_loss}, the learner will have low loss after $k$ updates, such that:
\begin{equation}
minimize_{\phi} \mathbb{E}_{\mathcal{T}_j}\left[{\mathcal{L}_{\mathcal{T}_j}\left(U^k_{\mathcal{T}_j}(\phi)\right)}\right],
\label{eq:metalearn}
\end{equation}
where $U^k_{\mathcal{T}_j}$ is the operator (e.g. corresponding to Adam optimizer or SGD) that updates $\phi$ using $k$ mini-batches from $\mathcal{D}^\train_j$.

\subsubsection{\textbf{Implementation Details}}
FS-1 sequentially samples few-shot tasks from the set of tasks $\mathcal{S}^\train$.
As summarized in Algorithm \ref{algo:fsl} and depicted in Figure \ref{fig:training}, the meta-learning procedure consists of $M$ meta-iterations. Each meta-iteration involves sampling $B$ $K$-shot tasks.
Each task, in turn, is solved using $k$ steps of gradient-based optimization, e.g. using stochastic gradient descent (SGD) or Adam \cite{kingma2015adam} -- this, in turn, involves randomly sampling mini-batches from the $K\times N$ instances in the task.
Each task is associated with a triplet loss defined over the valid triplets as described in Section \ref{ssec:loss}.

Given that each task has a varying number of instances owing to varying $N$, we set the number of iterations for each task to $k = \lfloor\frac{K\times N}{b}\rfloor\times e$, where $b$ is the mini-batch size and $e$ is the number of epochs. 
Therefore, instead of fixing the number of iterations $k$ for each sampled task, we fix the number of epochs $e$ across datasets, such that the network is trained to adapt quickly in a fixed number of epochs, as described later.
Also note that the number of triplets in each batch is significantly more than the number of unique time series in a mini-batch.

\begin{algorithm}
	\caption{Few-Shot UTSC Approach-1 (FS-1) \label{algo:fsl}}
	\begin{algorithmic}
		\State $\phi^0$: initial parameters of the ResNet
		\For{meta-iteration $i = 1,2,\dots, M$}
		\For{$j = 1,2,\dots, B$}
		\State \textit{Sample} a $K$-shot task $\mathcal{T}_j$
		\State \textit{Get} number of classes $N_j$ for task $\mathcal{T}_j$
		\State \textit{Set} $k = \lfloor\frac{K\times N_j}{b}\rfloor\times e$
		\State \textit{Compute} $\tilde \phi_j^i = U^k_{\mathcal{T}_j}(\phi^{i-1})$ using $k$ steps (mini-batches) of Adam to minimize loss $\mathcal{L}_{\mathcal{T}_j}$
		
		\EndFor
		\State Update $\phi^i = \phi^{i-1} + \epsilon \frac{1}{B}\sum_{j=1}^B( \tilde \phi_j^i - \phi^{i-1})$
		\EndFor
	\end{algorithmic}
\end{algorithm}

\begin{algorithm}
	\caption{Few-Shot UTSC Approach-2 (FS-2) \label{algo:tl}}
	\begin{algorithmic}
		\State $\phi^0$: initial parameters of the ResNet
		\For{iteration $i= 1,2,\dots,M$}
		\For{$j = 1,2,\dots, B$}
		\State \textit{Sample} a $K$-shot task $\mathcal{T}_j$
		\State \textit{Get} number of classes $N_j$ for task $\mathcal{T}_j$
		\State \textit{Set} $k = \lfloor\frac{K\times N_j}{b}\rfloor\times e$
		\State \textit{Compute} $ \phi^{i+j} = U^k_{\mathcal{T}_j}(\phi^{i+j-1})$ using $k_j$ steps (mini-batches) of SGD or Adam to minimize loss $\mathcal{L}_{\mathcal{T}_j}$
		\EndFor
		\EndFor
	\end{algorithmic}
\end{algorithm}

The filter weights of the ResNet are randomly initialized, e.g. via orthogonal initialization \cite{saxe2013exact}.
In the $i$th meta-iteration, ResNet for each of the $B$ tasks is initialized with $\phi^{i-1}$. 
Each task $\mathcal{T}_j$ with labeled data $\mathcal{D}^\train_j$ is solved by updating the parameters $\phi^{i-1}$ of the network $k$ ($=\lfloor\frac{K\times N_j}{b}\rfloor\times e$, where $N_j$ is number of classes in $\mathcal{T}_j$) times to obtain 
\begin{equation}
\tilde \phi_j^i = U^k_{\mathcal{T}_j}(\phi^{i-1}).
\label{eq:update}
\end{equation}

In practice, we use a batch version of the optimization problem in Equation \ref{eq:metalearn} and use a meta-batch of $B$ tasks to update $\phi$ as follows:
\begin{align}
\phi^{i} = \phi^{i-1} + \epsilon \frac{1}{B}\sum_{j=1}^B( \tilde \phi_j^{i} - \phi^{i-1}).
\label{eq:meta-update}
\end{align}

Note that $\tilde \phi_j - \phi$ with $k>1$ implies that $\phi$ is updated using the updated values $\tilde \phi_j$ obtained after solving $B$ tasks for $k$ iterations each. 
It is this particular way of updating $\phi$ by internally solving multiple tasks, that this algorithm is considered an example of gradient descent based meta-learning.
As shown in \cite{nichol2018first}, when performing multiple gradient updates as per Eqs. \ref{eq:update} and \ref{eq:meta-update}, i.e. having $k>1$ while solving few-shot tasks, then the expected update $\mathbb{E}_{\mathcal{T}_j}\big[{U^k_{\mathcal{T}_j}(\phi)}\big]$ is very different from taking a gradient step on the expected loss $\mathbb{E}_{\mathcal{T}_j}\big[\mathcal{L}_{\mathcal{T}_j}(\phi)\big]$, i.e. having $k=1$.
In fact, it is easy to note that the update of $\phi$ consists of terms from the second-and-higher derivatives of $\mathcal{L}_{\mathcal{T}_j}$ due to the presence of derivatives of $\mathcal{L}_{\mathcal{T}_j}$ in $\tilde \phi_j$.
Hence, the final solution using $k>1$ is significantly different from the one obtained using $k=1$.

\subsubsection{\textbf{Fine-tuning and inference in a test $K$-shot task}}
We denote the optimal parameters of ResNet after meta-training as $\phi^*$, and use this as initialization of target task-specific ResNet.
For any new $K$-shot $N$-way test task with labeled instances in $\mathcal{D}^\train$ and any test time series $\mathbf{x}^*$ taken from $\mathcal{D}^\test$, first $\phi^*$ is updated to $\tilde \phi$ using $\mathcal{D}^\train$. The embeddings for all the $N\times K$ samples in $\mathcal{D}^\train$ is compared to the embedding for $\mathbf{x}^*$ using 1NN classifier to get the class estimate.

\subsection{FS-2 \label{ssec:tl}}
As shown in Algorithm \ref{algo:tl}, FS-2 is a simpler variant of FS-1 where instead of updating the parameters $\phi$ by collectively using updated values from $B$ tasks, $\phi$ is continuously updated at each mini-batch irrespective of the task.
As a result, the network is trained for a few iterations on a task, and then the task is changed.
Unlike FS-1, FS-2 uses only the first-order derivatives of $\mathcal{L}_{\mathcal{T}_j}$.

\section{Experimental Evaluation\label{sec:exp}}
\subsection{Experimental Setup}
\subsubsection{\textbf{Sampling few-shot UTSC tasks}\label{sssec:sampling}}
We restrict the distribution of tasks to univariate TSC with a constraint on the maximum length of the time series such that $T\leq 512$.
We sample tasks from the publicly available UCR Archive of UTSC datasets \cite{UCRArchive}, where each dataset corresponds to a $N$-way multi-class classification task with number of classes $N$ and the length of time series $T$ varies across datasets. 
However, all the time series in any dataset are of same length. 
Each time series is $z$-normalized using the mean and standard deviation of all the points in the time series.

Out of the total of 65 datasets on UCR Archive with $T\leq 512$, we use 18 datasets to sample tasks for training meta-set $\mathcal{S}^\train$ and 6 datasets to sample tasks for the validation meta-set $\mathcal{S}^\valid$ (dataset level splits are same as in \cite{malhotra2017timenet}).
Any task in $\mathcal{S}^\train$ or $\mathcal{S}^\valid$ has $K$ randomly sampled time series for each of the $N$ classes in the dataset.
The remaining 41 datasets with length $T\leq512$ as listed in Table \ref{table:comparison} are used to create tasks for the testing meta-set.
As a result of this way of creating the training, validation and testing meta-sets, the classes in each meta-set are disjoint. However, the classes in the train and test sets of a task in a testing meta-set is, of course, the same.

\begin{figure}[h]
	\includegraphics[width=\columnwidth,trim={0cm, 1cm, 0cm, 1cm},clip]{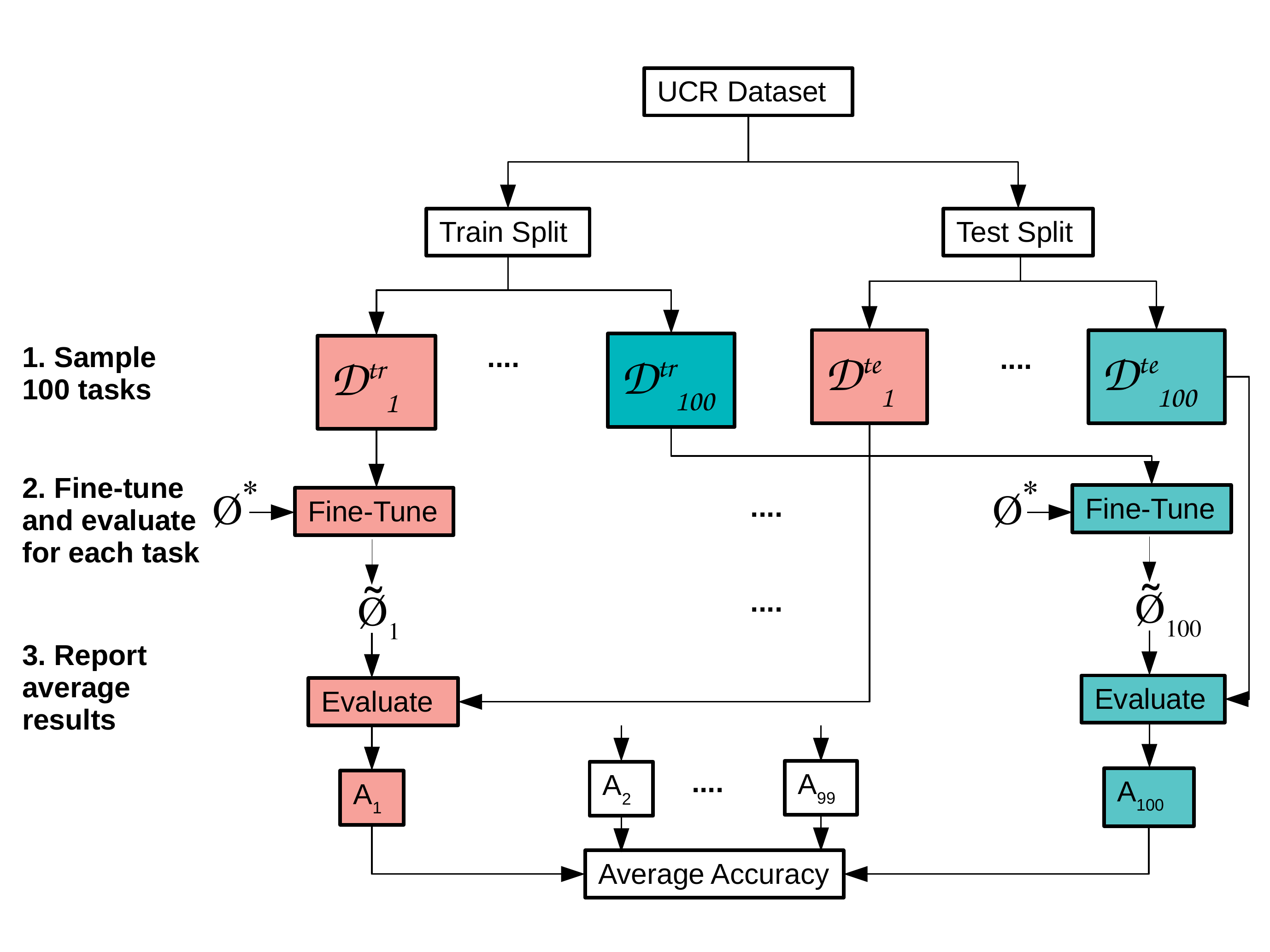}
	\caption{Evaluation protocol for FS-1 and FS-2 on a UCR dataset. For ResNet, $\phi$ is randomly initialized for each task. $A_j$ is the accuracy on $j$-th task.}
	\label{fig:exp_flow}
\end{figure}

Each dataset in UCR Archive is a $N$-way classification problem with an original train and test split.
As shown in Figure \ref{fig:exp_flow}, we sample 100 $K$-shot tasks from each of the 41 datasets. 
Each task (out of the 100) sampled from a dataset contains $K$ samples from each of the $N$ classes for $\mathcal{D}^\train$ and $K'$ samples from each of the $N$ classes for $\mathcal{D}^\test$ for each task are sampled from the respective original train and test split of the dataset\footnote{We also considered the original test split for each test task $\mathcal{D}^\test$ during evaluation. We obtained similar conclusions under this evaluation strategy as well, and hence, omit those results for brevity.}. 
The $K$ (or $K'$) samples for each class in $\mathcal{D}^\train$ (or $\mathcal{D}^\test$) are sampled uniformly from the entire set of samples of the respective class. 
While $\mathcal{D}^\train$ is used for fine-tuning $\phi^*$ to get $\tilde \phi$, $\mathcal{D}^\test$ is used to evaluate the updated task-specific model $\tilde \phi$.
(Note that while the class distribution in the original dataset may not be uniform, each $K$-shot task consists of equal number, i.e. $K$, samples per class.) 

\subsubsection{\textbf{Hyperparameters for FS-1 and FS-2}}
On the basis of initial experiments on a subset of the training meta-set, we use the ResNet architecture with $L=4$ layers and $m=165$ convolution filters per layer (33 filters each of length 4,8,16,32,64). 
We use Adam optimizer with a learning rate of 0.0001 for updating $\phi$ on each task while using $\epsilon=1$ in the meta-update step in Equation \ref{eq:meta-update}. 
FS-1 and FS-2 are trained for a total of $M=2000$ meta-iterations with meta-batch size of $B=5$, and mini-batch size $b=10$.
We trained FS-1 and FS-2 using $K=5$ and $10$ for the tasks in training meta-set while $K=5$ is used for validation and testing meta-sets. 
$K'=5$ across all experiments unless stated otherwise. 
We found the model with $K=10$ for tasks in training meta-set to be better based on average triplet loss on validation meta-set.
We use epochs $e=4$ for solving each task while training FS-1 and FS-2 models. 
The number of epochs $e'$ to be used while fine-tuning for tasks in testing meta-set is chosen from the range 1-100 based on average triplet loss on tasks in validation meta-set. We found $e'=16$ and 8 to be best for FS-1 and FS-2 models, respectively.
Therefore, $\phi^*$ is fine-tuned for $e'$ epochs for each task in testing meta-set.
For the triplet loss, we use $\alpha=0.5$.

\subsubsection{\textbf{Baselines Considered}}
For comparison, we consider following baseline classifiers each using 1NN as the final classifier over raw time series or extracted features\footnote{For DTW and BOSS, we use implementations as available at \url{http://www.timeseriesclassification.com/code.php}.}: 

\begin{enumerate}[leftmargin=*]
	\item \textbf{ED}: 1NN based on Euclidean distance is the simplest baseline considered, where time series of length $T$ is represented by a fixed-dimensional vector of the same length. (Note: For any given dataset and subsequent tasks sampled from it, the length $T$ is same across samples, and hence 1NN based on ED is applicable.)
	\item \textbf{DTW}: 1NN based on dynamic time warping (DTW) approach is one of the highly effective and strong baseline for UTSC \cite{bagnall2017great}. We use leave-one-out cross-validation on $\mathcal{D}^\train$ of each task to find the best warping window in the range $w={0.02T,0.04T,\ldots,T}$, where $w$ is the window length and $T$ is the time series length.  
	\item \textbf{BOSS}: Bag-of-SFA-Symbols \cite{schafer2015boss} is a state-of-the-art time series feature extraction technique that provides time series representations while being tolerant to noise. 
	BOSS provides a symbolic representation based on Symbolic Fourier Approximation (SFA) \cite{schafer2012sfa} on each fixed-length sliding window extracted from a time series while providing low pass filtering and quantization for noise reduction. 
	The hyper-parameters, i.e. \textit{wordLength} and \textit{normalization} are chosen based on leave-one-out cross validation over the ranges $\{8,10,12,14,16\}$ and $\{True,False\}$ respectively, while default values of remaining hyper-parameters is used. 
	1NN is applied on the extracted features for final classification decision.
	\item \textbf{ResNet}: Instead of using $\phi^*$ obtained via FS-1 or FS-2 as a starting point for fine-tuning, we consider a ResNet-based baseline where the model is trained from scratch for each task using triplet loss. The architecture is same as those used for FS-1 and FS-2 (also similar to state-of-the-art ResNet versions studied in \cite{wang2017time,fawaz2018deep,kashi2019convtimenet}). 
	Given that each task has a very small number of training samples and the parameters are to be trained from scratch, ResNet architectures are likely to be prone to overfitting despite batch normalization. To mitigate this issue, apart from the same network architecture as FS-1 and FS-2, we also consider smaller networks with smaller number of trainable parameters. More specifically, we considered four combinations resulting from number of layers $=\{\frac{L}{2},L\}$ and number of filters per layer $=\{\lfloor{\frac{m}{2}\rfloor},m\}$, where $L=4$ and $m=165$.
	We consider the model with best overall results amongst these four combinations as baseline, viz. number of layers = 2 and number of filters = 165.
	For fair comparison, each ResNet model is trained for 16 epochs\footnote{We also tried training till 32 epochs for ResNet and found insignificant improvement in results.} as for FS-1.
\end{enumerate}

\begin{figure*}
	\centering
	\begin{subfigure}[b]{0.24\textwidth}
		\includegraphics[width=\textwidth]{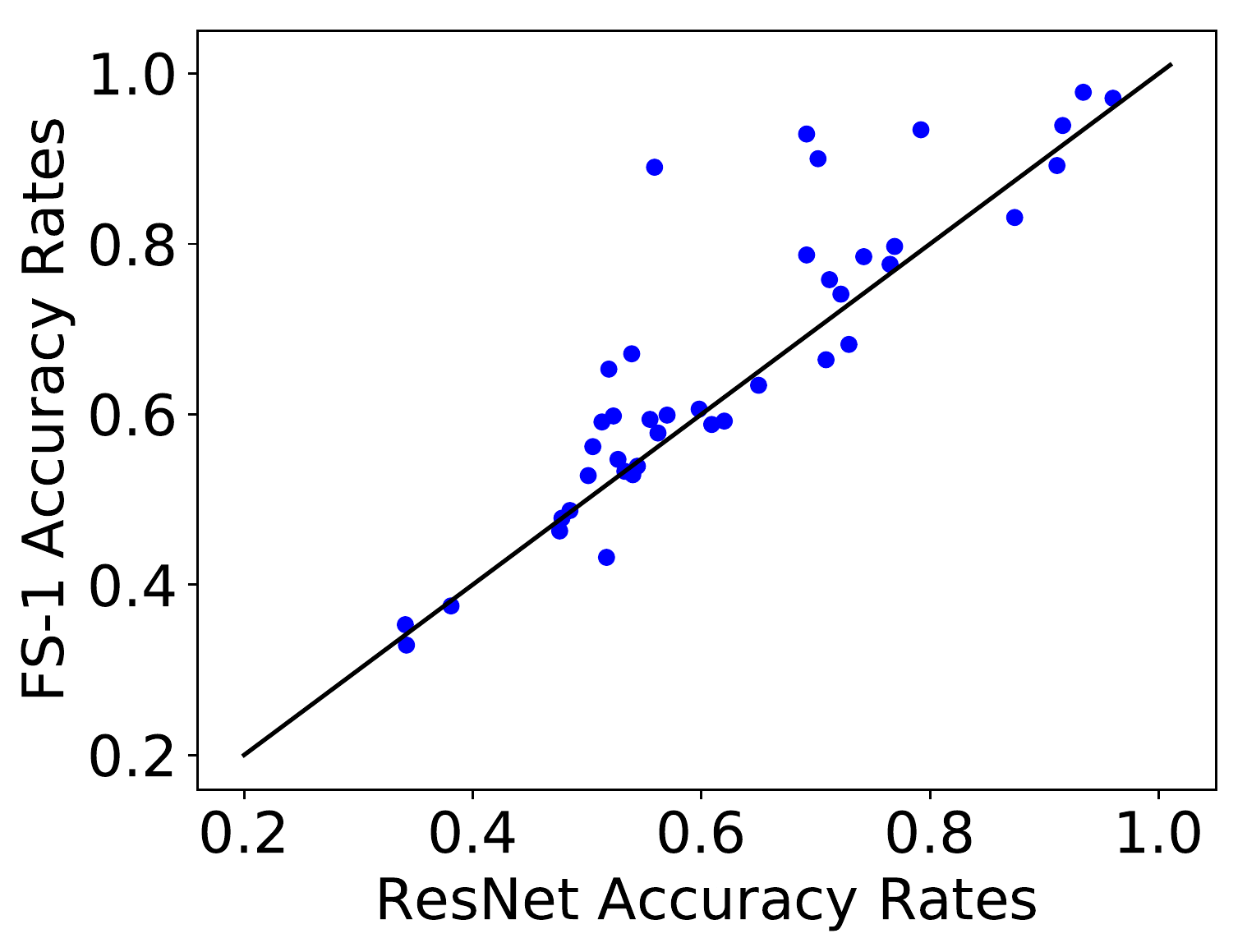}
		\caption{FS-1 vs ResNet}
		\label{fig:scat_fs1_fcn}
	\end{subfigure}
	\begin{subfigure}[b]{0.24\textwidth}
		\includegraphics[width=\textwidth]{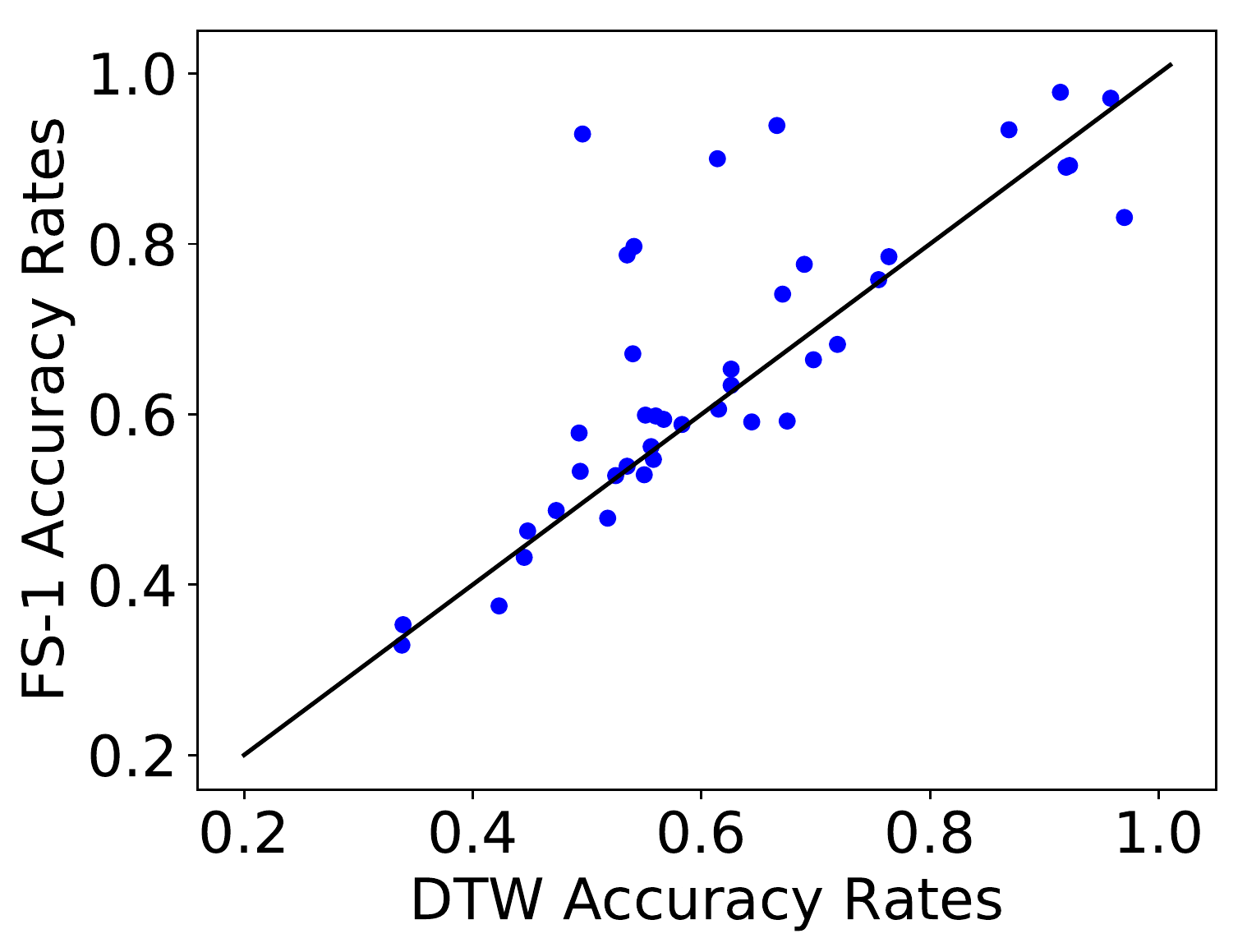}
		\caption{FS-1 vs DTW}
		\label{fig:scat_fs1_dtw}
	\end{subfigure}
	\begin{subfigure}[b]{0.24\textwidth}
		\includegraphics[width=\textwidth]{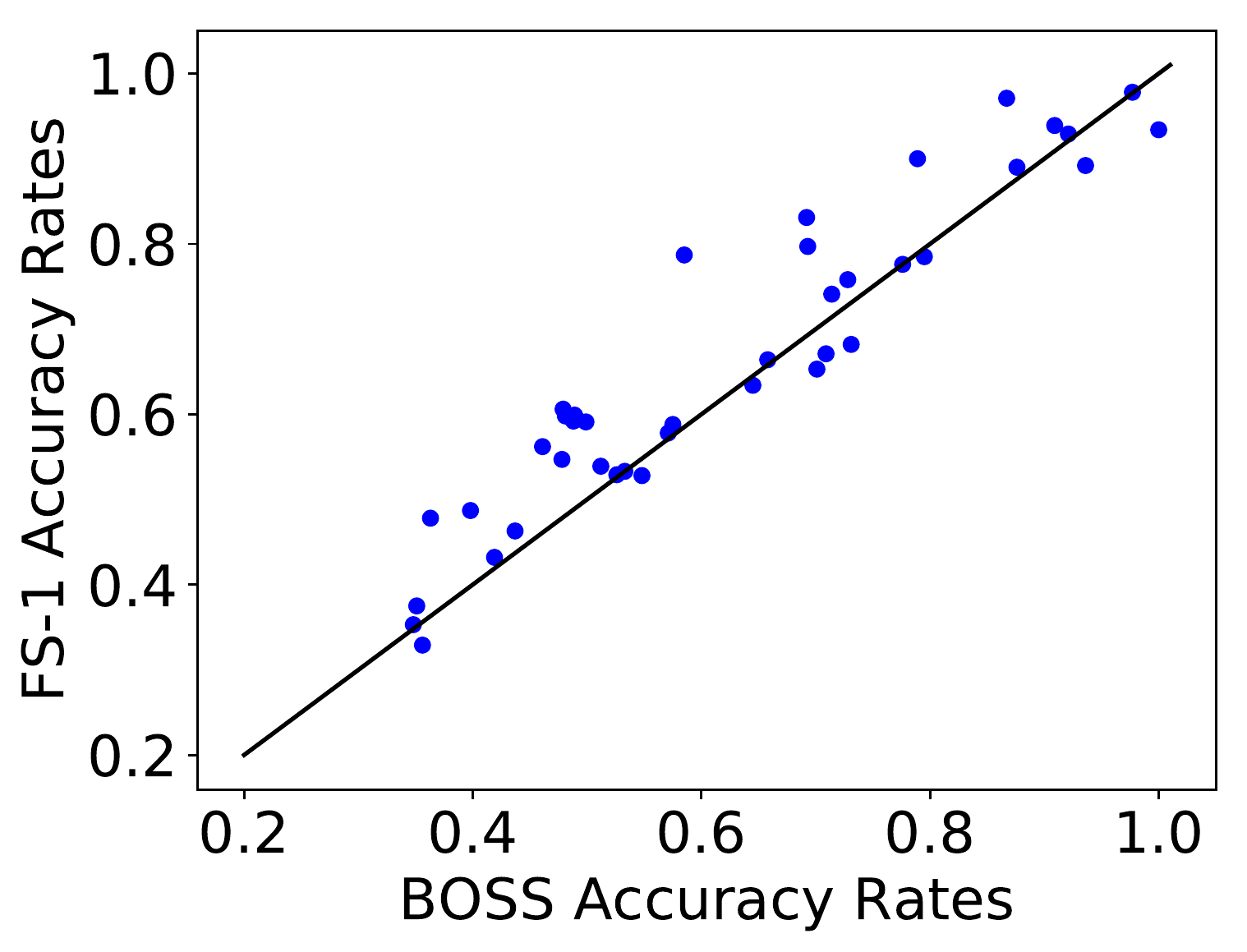}
		\caption{FS-1 vs BOSS}
		\label{fig:scat_fs1_boss}
	\end{subfigure}
	\begin{subfigure}[b]{0.24\textwidth}
		\includegraphics[width=\textwidth]{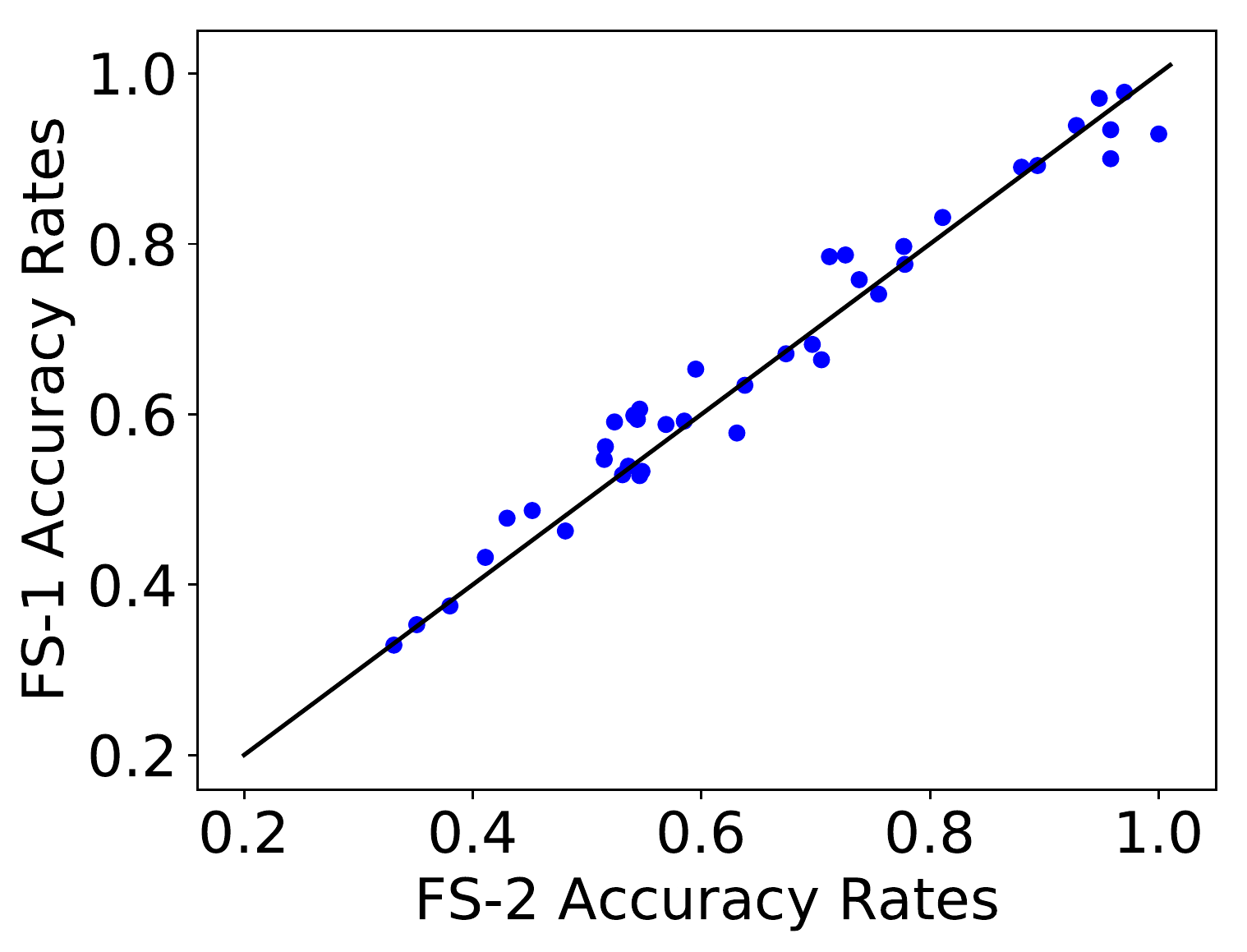}
		\caption{FS-1 vs FS-2}
		\label{fig:scat_fs1_fs2}
	\end{subfigure}
	\caption{Classification accuracy rates comparison for 5-shot UTSC. Each point in a scatter plot corresponds to a dataset.}\label{fig:scatter_plots}
\end{figure*}

\subsubsection{\textbf{Performance Metrics}}
Each task is evaluated using classification accuracy rate on the test set---inference is correct if the estimated label is same as the ground truth label. Each task consists of $K'\times N$ test samples: the performance results for each task equals the fraction of correctly classified test samples. 
Further, we follow the methodology from \cite{demvsar2006statistical,bagnall2017great} to compare the proposed approach with various baselines considered. 
For each dataset, we average the classification error results over 100 randomly sampled tasks (as described in Section \ref{sssec:sampling}).
To study the relative performance of the approaches over multiple data sets, we compare classifiers by ranks using the Friedman test and a post-hoc pairwise Nemenyi test. 
\begin{figure}[H]
	\includegraphics[width=0.85\columnwidth,trim={2cm, 0.4cm, 2cm, 0cm},clip]{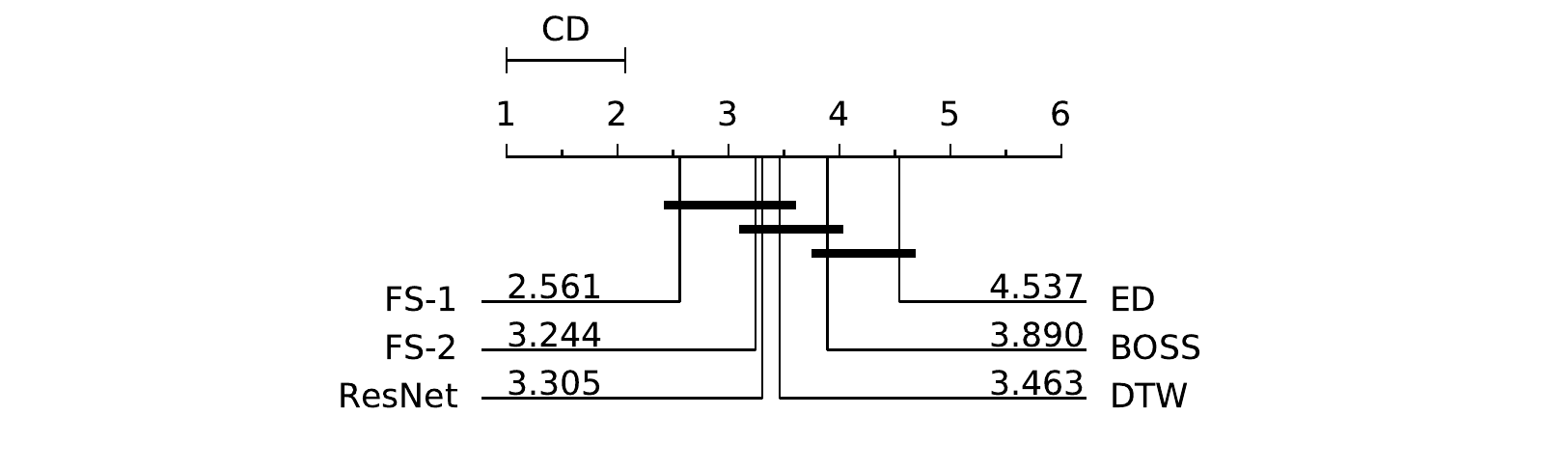}
	\caption{Critical Difference Diagram comparing ranks of few-shot learning approaches (FS-1 and FS-2) with other baselines for $K=5$ samples per class used for fine-tuning.}
	\label{fig:cd_benchmark}
\end{figure}

\begin{table*}[h]
	\footnotesize
	\centering
	\caption{Comparison of classification accuracy rates for 5-shot learning scenario. Best approach is marked in bold and second-best is underlined. N denotes the number of classes.}
	\begin{tabular}{|p{1.7cm}|c|c|c|c|c|c|c||p{1.9cm}|c|c|c|c|c|c|c|}
		\hline
		\textbf{Dataset Name} & \textbf{N} & \textbf{ED} & \textbf{DTW} &\textbf{BOSS} & \textbf{ResNet} & \textbf{\begin{tabular}[c]{@{}l@{}}FS-2\\(ours)\end{tabular}} & \textbf{\begin{tabular}[c]{@{}l@{}}FS-1\\(ours)\end{tabular}}
		&\textbf{Dataset Name} &  \textbf{N} & \textbf{ED} & \textbf{DTW} &\textbf{BOSS} & \textbf{ResNet} &
		\textbf{\begin{tabular}[c]{@{}l@{}}FS-2\\(ours)\end{tabular}} & \textbf{\begin{tabular}[c]{@{}l@{}}FS-1\\(ours)\end{tabular}} \\ \hline
		\textbf{50words} & 50 & 0.483 & \textbf{0.644} & 0.499 &0.513& 0.524 & \underline{0.591} & \textbf{InsectW.B.Sound} & 11 & \textbf{0.489} & 0.473 & 0.398 &0.485& 0.452 & \underline{0.487}\\
		\textbf{Adiac} & 37 & 0.538 & 0.540 & \textbf{0.709} &0.539& \underline{0.674} & 0.671 & \textbf{Meat} & 3 & \textbf{0.919} & \textbf{0.919} & 0.876 &0.559& 0.880 & \underline{0.890}\\
		\textbf{Beef} & 5 & 0.618 & 0.626 & \textbf{0.701} &0.519& 0.595 & \underline{0.653} & \textbf{MedicalImages} & 10 & 0.579 & \textbf{0.675} & 0.488 &\underline{0.620}& 0.585 & 0.592\\
		\textbf{BeetleFly} & 2 & 0.667 & 0.614 & 0.789 &0.702& \textbf{0.958} & \underline{0.900} & \textbf{Mid.Phal.O.A.G} & 3 & 0.529 & \textbf{0.558} & 0.478 &0.527& 0.515 & \underline{0.547}\\
		\textbf{BirdChicken} & 2 & 0.468 & 0.496 & 0.921 &0.692& \textbf{1.000} & \underline{0.929} & \textbf{Mid.Phal.O.C} & 2 & \textbf{0.563} & \underline{0.550} & 0.526 &0.540& 0.531 & 0.529\\
		\textbf{Chlor.Conc.} & 3 & 0.339 & 0.338 & \textbf{0.356} &\underline{0.342}& 0.331 & 0.329 & \textbf{Mid.Phal.TW} & 6 & 0.338 & 0.339 & 0.348 &0.341& \underline{0.351} & \textbf{0.353}\\
		\textbf{Coffee} & 2 & 0.920 & 0.914 & \underline{0.977} &0.934& 0.970 & \textbf{0.978} & \textbf{PhalangesO.C} & 2 & 0.532 & {0.535} & 0.512 &\textbf{0.544}& 0.536 & \underline{0.539}\\
		\textbf{Cricket\_X} & 12 & 0.348 & \underline{0.567} & 0.491 &0.555& 0.544  & \textbf{0.594} & \textbf{Prox.Phal.O.A.G} & 3 & 0.692 & 0.719 & \textbf{0.731} &\underline{0.729}& 0.697 & 0.682\\
		\textbf{Cricket\_Y} & 12 & 0.375 & \underline{0.556} & 0.461 &0.505& 0.516 & \textbf{0.562} & \textbf{Prox.Phal.O.C} & 2 & 0.633 & 0.626 & \underline{0.645} & \textbf{0.65} & 0.638 & 0.634\\
		\textbf{Cricket\_Z} & 12 & 0.357 & \underline{0.560} & 0.481 &0.523& 0.541 & \textbf{0.598} & \textbf{Prox.Phal.TW} & 6 & 0.427 & \underline{0.445} & 0.419 &\textbf{0.517}& 0.411 & 0.432\\
		\textbf{Dist.Phal.O.A.G} & 3 & \textbf{0.710} & 0.698 & 0.658 &\underline{0.709}& 0.705 & 0.664 & \textbf{Strawberry} & 2 & 0.682 & 0.671 & 0.714 &0.722& \textbf{0.755} & \underline{0.741}\\
		\textbf{Dist.Phal.O.C} & 2 & 0.571 & {0.583} & 0.575 &\textbf{0.609}& 0.569 & \underline{0.588} & \textbf{SwedishLeaf} & 15 & 0.599 & 0.690 & 0.776 &0.765& \textbf{0.778} & \underline{0.776}\\
		\textbf{Dist.Phal.TW} & 6 & 0.444 & 0.448 & 0.437 &\underline{0.476}& \textbf{0.481} & 0.463 & \textbf{synthetic\_control} & 6 & 0.736 & 0.958 & 0.867 &\underline{0.96}& 0.948 & \textbf{0.971}\\
		\textbf{ECG200} & 2 & \textbf{0.771} & 0.755 & 0.728 &0.712& 0.738 & \underline{0.758} & \textbf{Two\_Patterns} & 4 & 0.361 & \textbf{0.970} & 0.692 &\underline{0.874}& 0.811 & 0.831\\
		\textbf{ECG5000} & 5 & 0.524 & 0.494 & \underline{0.533} &\underline{0.533}& \textbf{0.548} & \underline{0.533} & \textbf{uWave\_X} & 8 & 0.591 & \textbf{0.615} & 0.479 &0.598& 0.546 & \underline{0.606}\\
		\textbf{ECGFiveDays} & 2 & 0.685 & 0.666 & 0.909 &0.916& \underline{0.928} & \textbf{0.939} & \textbf{uWave\_Y} & 8 & 0.504 & \textbf{0.518} & 0.363 &\underline{0.478}& 0.430 & \underline{0.478}\\
		\textbf{ElectricDevices} & 7 & 0.239 & \textbf{0.423} & 0.351 &\underline{0.381}& 0.380 & 0.375 & \textbf{uWave\_Z} & 8 & 0.536 & 0.551 & 0.489 &\underline{0.57}& 0.541 & \textbf{0.599}\\
		\textbf{FaceAll} & 14 & 0.545 & 0.764 & \textbf{0.795} &0.742& 0.712 & \underline{0.785} & \textbf{wafer} & 2 & \underline{0.922} & \underline{0.922} & \textbf{0.936} &0.911& 0.894 & 0.892\\
		\textbf{FaceFour} & 4 & 0.812 & 0.869 & \textbf{1.000} &0.792& \underline{0.958} & 0.934 & \textbf{Wine} & 2 & 0.496 & 0.493 & 0.571 &0.562& \textbf{0.631} & \underline{0.578}\\
		\textbf{FordA} & 2 & 0.561 & 0.541 & 0.693 &0.769& \underline{0.777} & \textbf{0.797} & \textbf{yoga} & 2 & 0.505 & 0.525 & \textbf{0.548} &0.501& \underline{0.546} & 0.528\\
		\cline{9-16}
		\textbf{FordB} & 2 & 0.515 & 0.535 & 0.585 &0.692& \underline{0.726} & \textbf{0.787}
		&\multicolumn{2}{l|}{\textbf{W/T/L of FS-1}} & 32/0/9 & 27/0/14 &
		30/2/9 &26/2/13& 24/0/17 & -\\
		\cline{1-8}
		\multicolumn{8}{c|}{}& \multicolumn{2}{l|}{\textbf{Mean Arithmetic Rank}} & 4.537 & 3.463 & 3.890 & 3.305 &
		3.244 & \textbf{2.561}\\
		\cline{9-16}	
	\end{tabular}
	\label{table:comparison}
\end{table*}

\begin{table}[H]
	\footnotesize
	\centering
	\caption{Comparison of various approaches in terms of ranks over classification accuracy rates on all the 4100 tasks from 41 datasets with varying $K$. Best approach is marked in bold and second-best is \underline{underlined}.}
	\begin{tabular}{|c|c|c|c|c|c|c|}\hline
		$\mathbf{K}$&\textbf{ED}&\textbf{DTW}&\textbf{BOSS}&\textbf{ResNet}&\textbf{FS-2}&\textbf{FS-1}\\\hline
		{2} & {4.232} & \underline{2.976} & {3.902} & {3.805} & {3.207} & \textbf{2.878}\\
		{5} & {4.537} & {3.463} & {3.890} & {3.305} & \underline{3.244} & \textbf{2.561}\\
		{10} & {4.573} & {3.476} & {3.646} & {3.683} & \underline{3.427} & \textbf{2.195}\\
		{20} & {4.439} & {3.354} & \underline{2.927} & {3.902} & {3.793} & \textbf{2.585}\\
		\hline
	\end{tabular}
	\label{table:varyingK}
\end{table}

\begin{table}[H]
	\footnotesize
	\centering
	\caption{Comparison of ranks across datasets with varying number of classes
		$N$ in 5-shot task and $n$ is the number of datasets.}
	\begin{tabular}{|c|c|c|c|c|c|c|c|}\hline
		$\mathbf{N}$&$n$&\textbf{ED}&\textbf{DTW}&\textbf{BOSS}&\textbf{ResNet}&\textbf{FS-2}&\textbf{FS-1}\\\hline
		{2-5} &24& {4.167} & 4.083 & 3.375 & 3.458 & \underline{3.042} & \textbf{2.875}\\
		{6-10} &9& 4.778 & \textbf{2.333} & 5.333 & \underline{2.389} & 3.778 & \underline{2.389}\\
		{$>$10} &8& 5.375 & 2.875 & 3.812 & {3.902} & 3.875 & \textbf{1.812}\\
		\hline
		Overall&41&{4.537} & {3.463} & {3.890} & {3.305} & \underline{3.244} & \textbf{2.561}\\\hline
	\end{tabular}
	\label{table:varyingN}
\end{table}

\subsection{Results and Observations}
\begin{itemize}[leftmargin=*]
	\item As shown in Figure \ref{fig:cd_benchmark}, we observe that FS-1 improves upon all the baselines considered for 5-shot tasks. The pairwise comparison of FS-1 with other baselines in Figure \ref{fig:scatter_plots} show significant gains in accuracies across many datasets.
	FS-1 has Win/Tie/Loss (W/T/L) counts of 26/2/13 when compared to the best non-few-shot-learning model, i.e. ResNet. 
	On 27/41 datasets, FS-1 is amongst the top-2 models. Refer Table \ref{table:comparison} for dataset-wise detailed results. Our approach FS-2 with a simpler update rule than FS-1 is the second best model but is very closely followed by the ResNet models trained from scratch. 
	\item To study the effect of number of training samples per class available in end task, we consider $K=\{2,5,10,20\}$ for $\mathcal{D}^\train$ (while $\mathcal{D}^\test$ remains the same with $K'=5$), and experiment under same protocol of 4100 tasks (with 100 tasks sampled from each of the 41 datasets). 
	As observed by ranks comparison in Table \ref{table:varyingK}, 
	\begin{itemize}
		\item FS-1 is the best performing model, especially for 5 and 10-shot scenarios with large gaps in ranks.
		\item When considering very small number of training samples per class, i.e. for $K=2$, we observe that FS-1 is still the best model although it is very closely followed by DTW. This is expected as given just two samples per class, it is very difficult to effectively learn any data distribution patterns, especially when the domain of the task is unseen while training. The fact that FS-1 and FS-2 still perform significantly better than ResNet models trained from scratch show the generic nature of filters learned in $\phi^*$. 
		As expected, data-intensive machine learning and deep learning models like BOSS and ResNet that are trained from scratch only on the target task data tend to overfit, and are even worse than DTW.
		\item For tasks with larger number of training samples per class, i.e. $K=20$, FS-1 is still the best algorithm. As expected, machine learning based state-of-the-art model BOSS performs better than other baselines when sufficient training samples are available and is closer to FS-1.		
	\end{itemize}
	\item To study the generalizability of FS-1 to varying $N$ as a result of leveraging triplet loss, we group the datasets based on $N$. As shown in Table \ref{table:varyingN}, we observe that FS-1 is consistently amongst the top-2 models across values of $N$. While FS-1 is significantly better than other algorithms for $2\leq N \leq 5$ and $N>10$, it is as good as the best algorithm DTW for $6\leq N\leq 9$.
\end{itemize}

\subsubsection{Importance of fine-tuning different layers in deep ResNet}
\begin{figure}[h]
	\includegraphics[width=0.85\columnwidth,trim={2cm, 0.4cm, 2cm, 0cm},clip]{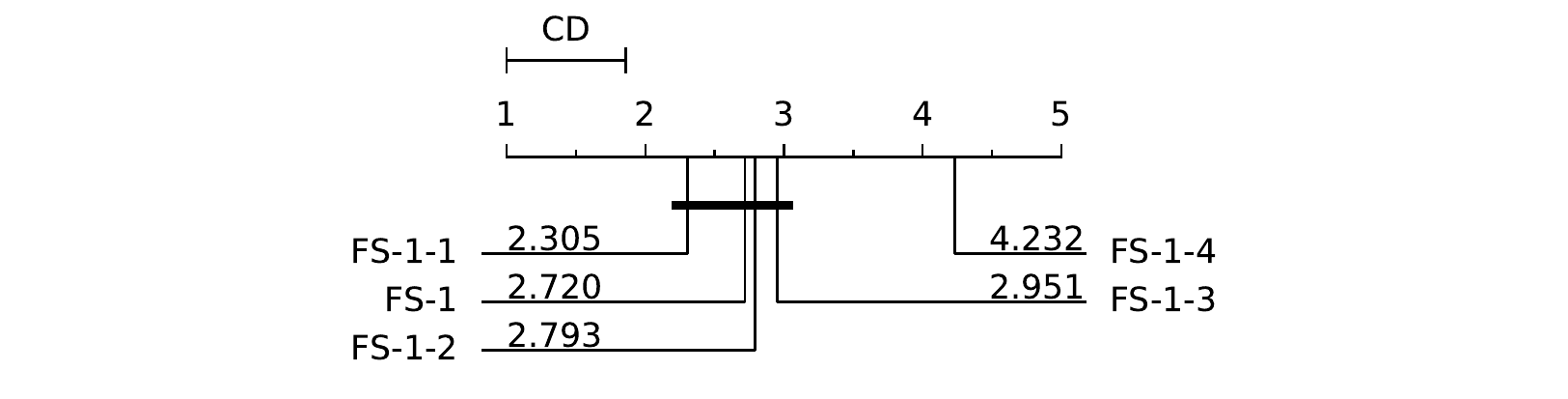}
	\caption{Effect of freezing parameters of different layers while fine-tuning for target few-shot task using FS-1.}
	\label{fig:cd_filters}
\end{figure}

We also study the importance of fine-tuning different convolutional layers of FS-1. 
We consider four variants FS-1-$l$ with $l=1,2,3,4$, where we freeze parameters of lowermost $l$ convolutional layers of the pre-trained model, while fine-tuning top $L-l$ layers only. 
From Figure \ref{fig:cd_filters}, we observe that FS-1-$1$, i.e. where the filter weights of only the first convolutional layer are frozen while those of all higher layers are fine-tuned, performs better than the default FS-1 model where all layers are fine-tuned. 
On the other hand, freezing higher layers as well (FS-1-2 and FS-1-3) or freezing all the layers (FS-1-4, i.e. no fine-tuning on target task) leads to significant drop in classification performance.
These results indicate that the first layer has learned generic features while being trained on diverse set of $K$-shot tasks and that the higher layers of the FS-1 model are important to quickly adapt to the target $K$-shot task.
\subsubsection{Few-shot learning to adapt to new classes for a given dataset}
\begin{table}[h]
	\footnotesize
	\centering
	\caption{Results on 5-shot 5-way classification tasks using dataset-specific pre-training.}
	\begin{tabular}{|c|c|c|c|c|c|c|}\hline
		$\mathbf{Dataset}$&\textbf{ED}&\textbf{DTW}&\textbf{BOSS}&\textbf{ResNet}&\textbf{FS-2}&\textbf{FS-1}\\\hline
		50Words & 0.614 & \textbf{0.812} & 0.713 & 0.733 & 0.719 & \underline{0.784}\\
		Adiac & 0.723 & 0.692 & 0.791 & 0.652 & \underline{0.808} & \textbf{0.827}\\
		ShapesAll & 0.854 & 0.897 & \underline{0.942} & 0.915 & 0.924 & \textbf{0.958}\\
		\hline
	\end{tabular}
	\label{tab:dataset_specific}
\end{table}
Apart from the above scenario where the UCR datasets used to sample tasks in training, validation and testing meta-sets are different, we also consider a scenario (similar to \cite{vinyals2016matching}) where there are a large number of classes within a TSC dataset, and the goal is to quickly adapt to a new set of classes given a model that has been pre-trained on another disjoint set of classes from the same dataset. 

We consider three datasets with large number of classes from the UCR Archive, namely, 50Words, Adiac and ShapesAll, containing 50, 37, and 60 classes, respectively. We use half of the classes (randomly chosen) to form the training meta-set, 1/4th of the classes for validation meta-set, and remaining 1/4th of the classes for testing meta-set. We train the FS-1 and FS-2 models on 5-shot 5-way TSC tasks from training meta-set for $M=50$ and $B=5$. 
We chose the best meta-iteration based on average triplet loss on the validation meta-set (also containing 5-shot 5-way classification tasks). 
Note that ED, DTW and BOSS are trained on the respective task from the testing meta-set only. Also, whenever number of samples for a class is less than 5, we take all samples for that class in all tasks. 
The average classification accuracy rates on 100 5-shot 5-way tasks from the testing meta-set are shown in Table \ref{tab:dataset_specific}. 
We observe that FS-1 outperforms other approaches indicating the ability to quickly generalize to new classes for a given domain.

\subsubsection{Non-few-shot learning scenario}
\begin{figure}[h]
	\centering
	\begin{subfigure}[b]{0.65\columnwidth}
		\includegraphics[width=1.2\textwidth,trim={2cm, 0.4cm, 2.2cm, 0cm},clip]{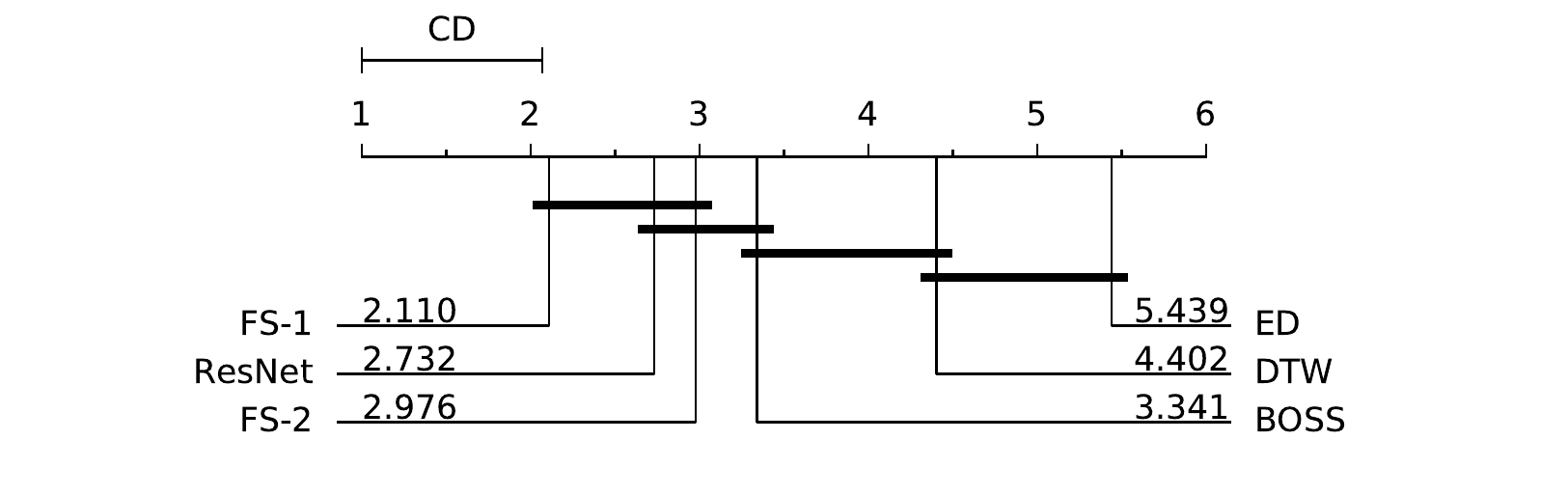}
		\caption{Critical Difference Diagram}
		\label{fig:cd_fulldata}	
	\end{subfigure}
	\begin{subfigure}[b]{0.65\columnwidth}
		\includegraphics[width=0.8\textwidth]{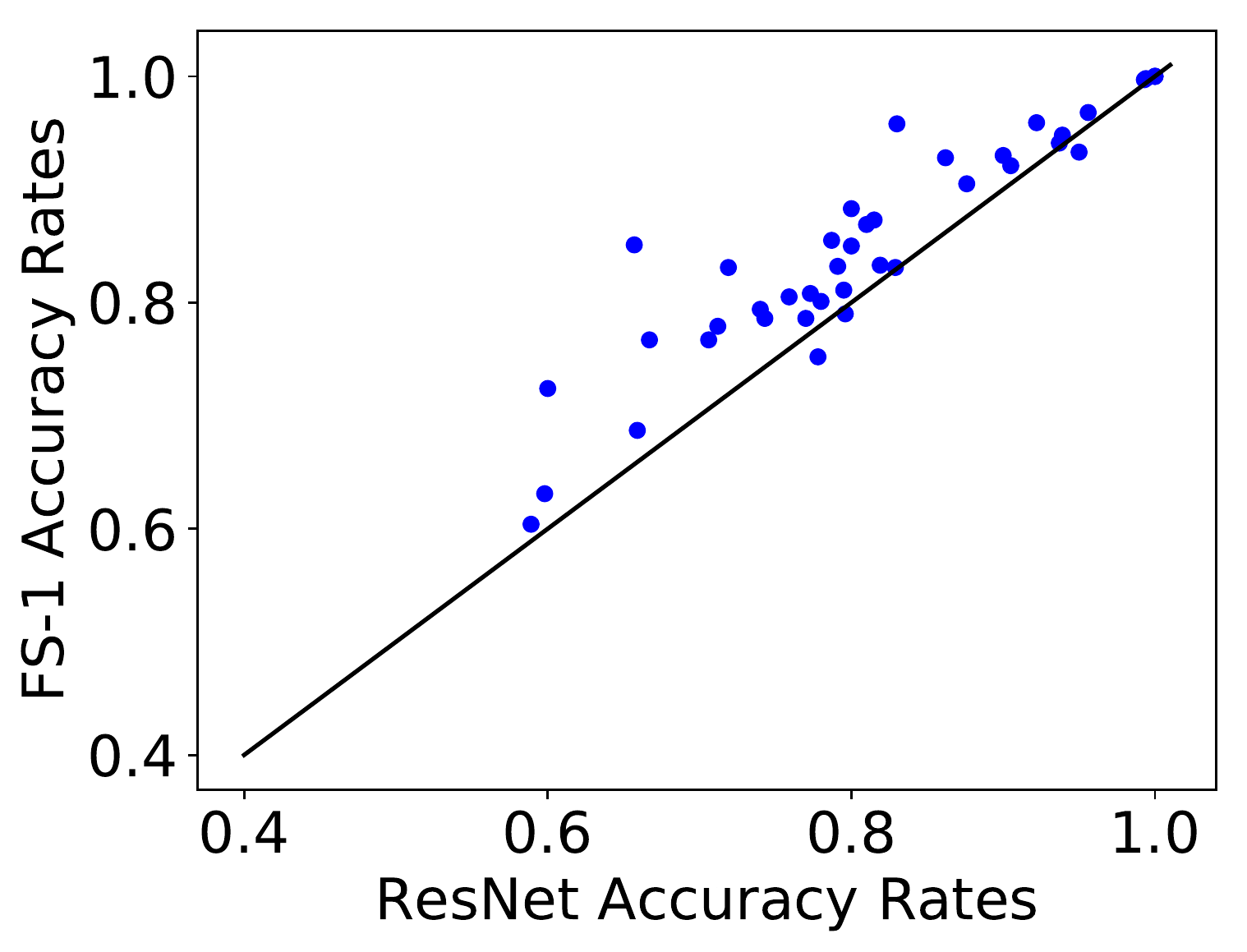}
		\caption{FS-1 vs second best method (ResNet)}		
		\label{fig:scat_fs1_ctnt}
	\end{subfigure}
	\caption{ Non-few-shot learning scenario using original train-test splits from UCR Archive.}
\end{figure}
We also evaluate FS-1 when sufficient labeled data is available for training, i.e. the standard non-few-shot learning scenario with original class distributions and train-test splits as provided in \cite{UCRArchive}. 
As shown in Figure \ref{fig:cd_fulldata}, we observe that the meta-learned FS-1 outperforms other approaches even in non-few-shot scenarios proving the benefit of meta-learning based initialization.
Furthermore, when compared to the results in Figure \ref{fig:cd_benchmark}, we observe increased performance gap between the deep learning approaches (FS-1, FS-2 and ResNet) and other approaches (BOSS, DTW, ED) due to availability of sufficient training data.
We provide scatter-plot comparison for FS-1 with second best approach ResNet in Figure \ref{fig:scat_fs1_ctnt} and omit other dataset-wise results for lack of space.

\section{Conclusion and Future Work\label{sec:conc}}
The ability to quickly adapt to any given time series classification task with a small number of labeled samples is an important task with several practical applications. 
We have proposed a meta-learning approach for few-shot time series classification (TSC).
It can also be seen as a data-efficient metric learning mechanism that leverages a pre-trained model.
We have shown that it is possible to train a model on few-shot tasks from diverse domains such that the model gathers an ability to quickly generalize and solve few-shot tasks from previously unseen domains.
By leveraging the triplet loss, we are able to generalize across classification tasks with different number of classes.

We hope that this work opens a promising direction for future research in meta-learning for time series modeling.
In this work, we have explored first-order meta-learning algorithms. In future, it would be interesting to explore more sophisticated meta-learning algorithms such as \cite{finn2017model,rusu2018meta,finn2018probabilistic} for the same.
A similar approach for time series forecasting will be interesting to explore as well. 

\bibliographystyle{ACM-Reference-Format}
\bibliography{BibTex/sensor_analytics,BibTex/esann17,BibTex/ijcnn19,BibTex/cikm19}

\end{document}